\documentclass[11pt,a4paper]{article}

\usepackage[a4paper,margin=2.5cm]{geometry}
\usepackage[T1]{fontenc}
\usepackage[utf8]{inputenc}
\usepackage{lmodern}
\usepackage{microtype}

\usepackage{authblk}

\usepackage[numbers,sort&compress]{natbib}
\providecommand{\citeauthoryear}[3]{#1 (#3)}

\usepackage{amsmath}
\usepackage{amssymb}
\usepackage{amsfonts}
\usepackage{bm}
\usepackage{dsfont}

\usepackage{multirow}
\usepackage{graphicx}
\usepackage{booktabs}
\usepackage{xcolor}
\usepackage{siunitx}
\usepackage{caption}
\usepackage{subcaption}
\usepackage{pgfplots}
\pgfplotsset{compat=1.18}
\usepackage{tikz}
\usetikzlibrary{decorations.pathmorphing,arrows.meta,calc,positioning,patterns,backgrounds}

\usepackage{xspace}
\usepackage{etoolbox}
\usepackage[colorlinks=true,citecolor=blue!70!black,linkcolor=blue!70!black,urlcolor=blue!70!black]{hyperref}

\graphicspath{{./}{figures_rev2/}}

\def\tsc#1{\csdef{#1}{\textsc{\lowercase{#1}}\xspace}}
\tsc{WGM}
\tsc{QE}

\title{\textbf{Uncertainty-aware damage identification in short-span bridges via physics-informed variational autoencoder}}

\author[1]{Ana Fernandez-Navamuel}
\author[2]{\'{A}ngel J. Omella}
\author[3]{Diego Zamora-Sanchez}
\author[4,1]{David Pardo}

\affil[1]{Basque Center for Applied Mathematics (BCAM), Bilbao, Spain}
\affil[2]{Departamento de Matem\'{a}tica Aplicada, Universidad de Zaragoza/IUMA, Zaragoza, Spain}
\affil[3]{TECNALIA, Basque Research and Technology Alliance (BRTA), Parque Cient\'{\i}fico y Tecnol\'{o}gico de Bizkaia, Astondo bidea, Edificio 700, E-48160 Derio, Spain}
\affil[4]{Department of Mathematics, University of the Basque Country (UPV/EHU), Spain}

\date{\today}

\begin{document}

\maketitle

\begin{abstract}
\noindent Vibration-based damage identification in civil infrastructure is a challenging, ill-posed inverse problem due to measurement noise, sparse sensor arrays, and modeling errors. While deep learning is powerful for system identification, deterministic approaches lack reliable uncertainty quantification and can yield physically inconsistent results. This work proposes a probabilistic Scientific Machine Learning (SciML) framework for structural health monitoring (SHM): a physics-informed Gaussian copula variational autoencoder (PI-GCVAE). The methodology introduces two main contributions. First, a differentiable numerical eigenvalue solver is embedded directly into the variational autoencoder as its decoder, which replaces data-driven surrogates, ensures that every posterior sample satisfies the governing equations of structural dynamics, and reduces the number of trainable parameters. Second, a Gaussian copula describes the posterior distribution of the stiffness factors, capturing the cross-correlations between structural components with fewer parameters than Gaussian mixtures. The framework is trained without damage labels and provides the posterior distribution in real time. We validate the method numerically on a short-span bridge supported on elastic bearings, whose response is dominated by vertical bending, accounting for modeling errors, measurement noise, and a finer finite element discretization for testing. The method estimates the posterior distribution of the damage condition, i.e., the stiffness reduction factors of the deck zones and bearings, and these posteriors closely match the exact ones computed with sequential Monte Carlo sampling. The results reveal that the proposed methodology offers a robust and interpretable decision-making tool, able not only to indicate the presence and location of damage, but also to inform on the uncertainty associated with the diagnosis, for instance when light damage is masked by the effect of noise or when the measured modal properties present limited sensitivity to damage.
\end{abstract}

\section{Introduction}
\label{sec:intro}
Vibration-based structural health monitoring (SHM) is increasingly applied to assess the condition of large-scale operative systems \cite{Jayasundara2020,Avci2021_VSHMprogress}.
It addresses an inverse problem known as system identification \cite{ChatziPapadi2016identification}, which aims to discover the presence of damage from measurements of a system's dynamic response \cite{VSHM_Liu2021,Jayasundara2020,Casas2017}.
In the context of civil engineering, structural damage is defined as any reduction in the system's performance capabilities resulting from a local or distributed decrease in its stiffness properties \cite{Friswell2007b}.
The measured response often comes from acceleration sensors sparsely distributed along the structure. However, since raw acceleration signals are highly sensitive to environmental and operational conditions, a standard practice is to process them using operational modal analysis (OMA) techniques \cite{Raineri_eigenmodes,AutomatedOMA2021,MACEC_Reynders2014}.
The resulting modal properties — natural frequencies and mode shapes — are isolated from external variability, making them highly suitable inputs for damage identification.

Historically, solving the inverse problem of damage identification has been addressed via two main approaches: non-parametric (signal-based) and parametric (model-based) methods \cite{fan2011vibration}.
Signal-based methods, such as those utilizing wavelet transforms \cite{bayissa2008vibration}, Hilbert-Huang transforms \cite{roveri2012damage}, or statistical process control (SPC) \cite{fugate2001vibration} and pattern recognition \cite{bishop2006pattern}, detect anomalies directly from raw or processed sensor data without requiring a mathematical model of the structure.
While computationally efficient, these methods often struggle to provide a precise quantification of damage severity or location. Conversely, parametric model-based methods, particularly finite element model updating (FEMU) \cite{Teughels2002Damage,FEMU2004}, calibrate a structural numerical model until its output matches experimental observations, e.g., to identify the progressive damage~\cite{teughels2004z24} or to estimate the stiffness of the deck and elastomeric bearings~\cite{elborgi2005modal}.
Although FEMU provides physical interpretability, it is often computationally prohibitive for real-time monitoring.

A critical aspect of the inverse problem is nonuniqueness, which leads to multiple damage scenarios yielding the same observable response. This phenomenon owes to the inherent uncertainties arising mainly from (i) model uncertainty, (ii) measurement error and limitations, and (iii) environmental and operational variability \cite{liu2017efficient,simoen2015dealing}. Bayesian FEMU approaches have been proposed to tackle ill-posedness in structural damage identification \cite{simoen2015dealing,Giagopoulos2019_BayesFEMU}, and particularly to identify damage in bridges using modal data \cite{behmanesh2015probabilistic,zhou2022vibration,huang2017hierarchical}.
However, all these approaches employ Markov-chain Monte Carlo (MCMC) to ``blindly'' draw samples from a proposal posterior distribution throughout the updating process, demanding a massive number of forward simulations to achieve convergence for each new set of measurements \cite{ching2007transitional,ma2025copula}.

Nowadays, due to its real-time inference capabilities, damage identification is predominantly tackled using artificial intelligence (AI), and particularly deep neural networks (DNNs) \cite{BigDataSHM2020,zhang2022review}.
In a supervised learning scheme for bridge damage assessment, the DNN establishes a mapping between the modal properties (i.e., natural frequencies and mode shapes) to a set of parameters describing the target damage condition (e.g., material properties or boundary conditions) \cite{lee2005neural,Jayasundara2020,AFN_SupervisedSHM,AFN_SupervisedEOCs,wang2026data}.
According to this scheme, the loss function measures the discrepancy between the estimated and the true system properties.
In our previous work~\cite{AFN_SupervisedSHM}, we applied a deterministic strategy, parameterizing the damage condition using two features: severity and location.
In~\cite{AFN_SupervisedEOCs}, we expanded the synthetic training database to various representative environmental and operational conditions, thus covering most of the external variability.
These deterministic methods provide point estimates of the damage condition, without a measure of its reliability~\cite{sun2023deep,wang2026data}.
However, when attempting to identify early damage in operating civil engineering structures, this assessment is often overconfident due to the inherent ill-posedness, caused by
$(i)$ discrepancies between the target physical system and the computational parameterization used to generate the training data \cite{Sanayei2004}, $(ii)$ masking effect of noise and observation limitations~\cite{simoen2015dealing}, and $(iii)$ environmental variability \cite{ChatziPapadi2016identification,simoen2015dealing,whiteman2024convolutional}.

Autoencoder architectures use an encoder neural network (NN) to approximate the inverse problem and a decoder that acts as a regularizing surrogate that aids in finding consistent solutions by satisfying the forward problem \cite{AEfoundations1988,peng2020solving,tarantola2005inverse,Pardo_rescaling}.
Variational autoencoders (VAEs) have consolidated as a promising solution for robust damage identification \cite{pinheiro2021variational,wei2025parameterVAE,haciefendiouglu2026dual,liu2022uncertainty,Rodriguez2023,Navamuel_wes2025,bacsa2025structural}. VAEs map input measurements to a parametric probability density function (PDF) that describes the posterior distribution of the system parameters \cite{BayesInverse_Goh2021,VAE_tutorial}.

A critical limitation of standard VAEs is their reliance on diagonal-covariance Gaussian distributions, which fail to capture cross-correlations between latent variables. While this assumption significantly reduces computational complexity and enables efficient training, it comes at the cost of neglecting the underlying physics-driven dependencies between adjacent structural elements, which are essential in structural damage assessment.
Although more expressive variational families, such as Gaussian mixtures \cite{Navamuel_wes2025} or those based on normalizing flows \cite{papamakarios2021normalizing}, could in principle capture such complex dependencies, they incur prohibitive computational cost and training complexity.
A proof-of-concept of a Gaussian copula VAE approach was proposed to identify damage in a simplified two-dimensional problem~\cite{Navamuel2026copula2d}.
Copulas are functions that describe the dependence among interrelated random variables by disentangling the joint distribution and the independent marginals of each individual variable \cite{nelsen2003properties,joe2014dependence}.
However, scaling this approach to handle high-dimensional dependencies remains an open challenge.

An important deficiency of purely data-driven (variational) autoencoders lies in the lack of interpretability of the estimated outcomes, mainly when dealing with ill-posed problems~\cite{Navamuel_wes2025}.
Scientific machine learning (SciML) offers a hybrid strategy that embeds physical domain knowledge and governing equations directly into the network \cite{okazaki2025SciML}.
Physics-informed neural networks (PINNs) \cite{Karniadakis2019PhysicsInf,mo2025explainable} have become the most popular SciML approach, which uses automatic differentiation to solve partial differential equations \cite{koune2025adversarial,almanstotter2025pinnverse,teloli2025physics,zhang2025novel,hu2023structural}.
In the context of SHM, a PINN was embedded within a VAE architecture to expand sparse measurements into full-field data for damage identification in small metal plate structures~\cite{kulkarni2026full}.

Since using raw time histories is computationally inefficient and highly susceptible to noise and environmental variability, it is desirable to operate in the modal domain (OMA-extracted features). However, embedding the generalized eigenvalue problem into a neural network architecture has not been explored in SHM.

In this work, we propose a novel, probabilistic SciML approach to solve SHM inverse problems under uncertainty using limited modal data.
We extend the applicability of the Gaussian copula-based VAE to a cross-correlated multi-dimensional latent space to identify damage in a numerical model of a short-span bridge supported on elastic bearings.
One key advantage of this approach is the exact computation of the forward operator without increasing the search space complexity.
Rather than using a PINN to infer the entire displacement field, we directly update the modal properties based on estimated element stiffness matrices.
While a previous work incorporated numerical solvers as dynamic cells for time-domain state prediction, the authors found no prior work that embedded a generalized eigenvalue method to solve SHM inverse problems in the modal domain~\cite{eshkevari2021dynnet}.
We denote our method as physics-informed Gaussian copula VAE (PI-GCVAE).

Our proposed architecture comprises three main components: \textit{(a)} an encoder NN that maps measured modal properties to the parameters describing the posterior distribution of the stiffness reduction factors; \textit{(b)} a sampling layer that builds the parametric posterior PDF using a Gaussian copula and draws samples from it, accounting for the dependencies of adjacent structural components; and \textit{(c)} a differentiable numerical eigenvalue solver (decoder) that replaces the traditional NN decoder, enabling the exact computation of the forward problem to deliver the estimated modal properties.

We train and evaluate the proposed methodology using a synthetic dataset from a numerical parameterization of a short-span reinforced concrete bridge supported on elastic bearings.
Although the numerical model provides access to the full modal matrix, we retain only the lowest-order mode shapes.
This constraint exists in real SHM campaigns, where the OMA applies to sparse sensor data, reliably capturing only the fundamental, highly energetic mode shapes.
With these specifications, the employed synthetic dataset constitutes a reliable and fair representation of the actual measurements obtained in operating bridges.

The main contributions of this work are twofold:
\begin{itemize}
	\item \textbf{Embedded differentiable eigenvalue solver:} We incorporate a numerical method to solve the forward problem, embedded directly within the VAE architecture. It overcomes the limitations of standard autoencoders that rely purely on data to approximate the forward mapping (decoder) \cite{Navamuel_wes2025,bacsa2025structural}, eliminating surrogate error and reducing the number of trainable parameters while bypassing the training difficulties of PINNs \cite{eshkevari2021dynnet}.
	\item \textbf{Gaussian copula posterior multivariate latent space:} We implement a parameter-efficient posterior PDF based on a Gaussian copula to describe the latent space uncertainty. This approach captures the physical dependencies between the stiffness reduction parameters using far fewer parameters than Gaussian mixture models \cite{Navamuel_wes2025}, extending the achievements of \cite{Navamuel2026copula2d} to high-dimensional cross-correlated multivariate spaces.
\end{itemize}

The proposed methodology applies to structures whose dynamic response is described by a linear finite element parameterization, under two main assumptions.
First, damage remains within the linear elastic regime, i.e., we target early-stage damage, for which the modal properties follow from the generalized eigenvalue problem.
Second, the identified modes represent the predominant response of the structure.
In short-span bridges, such as reinforced concrete decks on elastomeric bearings, vertical bending is the predominant response, which enables a one-dimensional model with decoupled vertical bending modes~\cite{moliner2017dynamic}.
Transversal or torsional modes can also be included, although modal coupling may hinder the identification.

Within these assumptions, the method can be generalized as it only requires that the stiffness matrix is a weighted sum of matrices linked to the unknown parameters.
Hence, it can be applied to other structures and damage parameterizations (e.g., continuous bridges with intermediate bearings, slender structures such as wind turbines or truss frames) by extending the latent variables.

Although not experimentally validated, the methodology is oriented to its application in real practice.
It uses the data of a standard vibration monitoring system, i.e., the first natural frequencies and mode shapes obtained from few raw acceleration sensors, which are the most accurately identified modal properties.
It accounts for the main sources of uncertainty of real applications (i.e., measurement noise, and modeling errors), although environmental variability is considered future work within the experimental validation.
Once trained, it provides in real time, for each component, the posterior probability distribution given the observed data, and the probability of damage, so that a reliable decision can be taken with a known confidence level, preventing overconfident assessment.

\section{Problem formulation}
\label{sec:problem_formulation}
\subsection{Finite element formulation for a short-span bridge on elastic bearings}
\label{subsec:Matrix_form}
Let us assume a short-span bridge of length $L$ governed by its motion partial differential equation (PDE) under the Euler-Bernoulli beam theory \cite{przemieniecki1968}:
\begin{equation}
	EI \frac{\partial^4 \text{w}(x, t)}{\partial x^4} + \rho A \frac{\partial^2 \text{w}(x, t)}{\partial t^2} = p(x, t),
	\label{eq:PDE_formula}
\end{equation}
where $\text{w}(x, t)$ represents the transverse deflection field as a function of position ($x$) and time ($t$); the geometry and material properties are described by the constants $E$ (material elastic modulus), $I$ (cross-sectional second moment of inertia), $\rho$ (material density), and $A$ (cross-sectional area); we consider a distributed load per unit length denoted as $ p(x, t)$.
To obtain a unique solution for Eq.~\eqref{eq:PDE_formula}, we define the boundary conditions of a beam resting on elastic bearings at both ends ($x=0$ and $x=L$), represented by vertical springs of stiffness $K_{v}$, with free rotations: null bending moment and shear force balanced by the reaction of the springs \cite{moliner2017dynamic,erduran2022effect}:
\begin{equation}
	\begin{aligned}
		EI \frac{\partial^2 w(0, t)}{\partial x^2} &= 0, \quad & EI \frac{\partial^3 w(0, t)}{\partial x^3} &= -K_{v}\, w(0, t), \\
		EI \frac{\partial^2 w(L, t)}{\partial x^2} &= 0, \quad & EI \frac{\partial^3 w(L, t)}{\partial x^3} &= K_{v}\, w(L, t).
	\end{aligned}
	\label{eq:boundary_conditions}
\end{equation}
For the initial condition, we assume that the bridge starts from a rest condition, such that:
\begin{equation}
	w(x, 0) = 0, \quad \frac{\partial w(x, 0)}{\partial t} = 0.
	\label{eq:initial_conditions}
\end{equation}

We discretize the system using the finite element (FE) method to find a solution to the PDE.
We consider $n_{\text{el}}$ one-dimensional linear elastic elements to parameterize the structural behavior of the bridge.
In this work, we focus on the vertical bending displacements.
Thus, we neglect longitudinal and transversal movements.
The length of each element is $L_{e} = \frac{L}{n_{\text{el}}}$.
We now introduce $n_n =  n_{\text{el}} +1$ nodes. Each node is located at positions $x = \{ 0, L_{e}, 2 L_{e},..., L\}$ and has two associated degrees of freedom (DOFs): the vertical displacement $u(x,t)$ and the rotation $\xi(x,t)$ \cite{przemieniecki1968}.

By applying the principle of virtual work, the internal strain energy and kinetic energy of the element are transformed into the standard element stiffness matrix $\mathbf{K}_{e}$ and mass matrix $\mathbf{M}_{e}$, respectively \cite{pepper2005finite}.
We build the global matrices by assembling all the element matrices.
Finally, we impose the boundary conditions. Here, we add the stiffness $K_{v}$ of the bearing springs to the vertical DOFs of the end nodes, resulting in a total number of DOFs $n_{\text{dof}} = 2 \times n_{n}$.
We denote by $\mathcal{A}(\cdot)$ the function that assembles the element matrices and the bearing springs to form the $n_{\text{dof}}\times n_{\text{dof}}$ global matrices.
Then, we denote $\mathbf{K}_{\text{G}} = \mathcal{A}(\{\mathbf{K}_{e}\}_{e = 1}^{n_{\text{el}}})$ and $\mathbf{M}_{\text{G}} =  \mathcal{A}(\{\mathbf{M}_{e}\}_{e = 1}^{n_{\text{el}}})$ the global stiffness and mass matrices of the undamaged (baseline) structure, respectively.

\subsection{Damage parametrization and the generalized eigenvalue problem}
\label{sec:Parametrization}
We parameterize damage through reduction factors applied to the baseline stiffness of two types of structural components: $n_z$ damage zones along the deck, each comprising one or several consecutive elements, and $n_b$ bearings, represented by vertical springs.
Let $z_{d} \in [z_{\text{lb}}, z_{\text{ub}}]$ be a dimensionless reduction factor for the $d$-th component, where $z_{\text{ub}} = 1.0$ indicates the healthy state, $z_{d} < 1.0$ indicates damage, and $z_{\text{lb}} > 0$ represents the lower stiffness reduction considered in the analysis.
The global stiffness matrix is assembled from the element matrices, each one scaled by the stiffness factor of its component:
\begin{equation}
	\hat{\mathbf{K}}_{\text{G}}(\mathbf{z})=\sum_{k=1}^{n_z} z_k\sum_{e\in\mathcal{Z}_k}\mathbf{L}_e^{\top}\mathbf{K}_e\,\mathbf{L}_e
	+\sum_{r=1}^{n_b} z_{n_z+r}\,K_{v}\,\bm{\ell}_r\bm{\ell}_r^{\top},
	\label{eq:Global_Stiffness}
\end{equation}
where $\mathbf{L}_e$ is the Boolean connectivity matrix that extracts the DOFs of element $e$ from the global DOF vector, so that $\mathbf{L}_e^{\top}\mathbf{K}_e\mathbf{L}_e$ is the element matrix expanded to the global dimensions; $\mathcal{Z}_k$ is the set of elements of the $k$-th deck zone; and $\bm{\ell}_r$ is the unit vector that selects the vertical DOF of the $r$-th bearing.
We gather these factors into a damage condition vector $\mathbf{z} = \{z_{1}, \dots, z_{D}\}^\top \in \mathcal{S}$, with $D=n_z+n_b$ and support $\mathcal{S} = [z_{\text{lb}}, z_{\text{ub}}]^{D}$.
In this work, $n_z=5$ deck zones and $n_b=2$ bearings (left and right, with stiffness factors $z_{b,L}\equiv z_6$ and $z_{b,R}\equiv z_7$), so $D=7$.
We represent the damage as a reduction of the effective bending stiffness of each zone \cite{Teughels2002Damage,moliner2017dynamic}.
The stiffness factor of a bearing represents a loss of vertical support stiffness, e.g., due to the deterioration of the elastomer or of its bedding.
In this work, we assume that the mass matrix remains invariant after the occurrence of damage.

The dynamic characteristics of the damaged system are dictated by the generalized eigenvalue problem.
Assuming undamped free vibration within the linear elasticity regime, the problem is formulated as:
\begin{equation}
	\hat{\mathbf{K}}_{\text{G}}(\mathbf{z}) \bm{\Phi} = \mathbf{M}_{\text{G}} \bm{\Phi} \bm{\Lambda},
	\label{eq:Eigenvalue_problem}
\end{equation}
where $\bm{\Lambda} = \text{diag}(\omega_1^2, \dots, \omega_{n_{\text{m}}}^2)$ is a diagonal matrix containing the squared angular frequencies and $\bm{\Phi} = [\bm{\phi}_1, \dots, \bm{\phi}_{n_{\text{m}}}]$ contains the first $n_{m}$ mass-normalized mode shapes.
The natural frequencies are given by $f_i = \omega_i / (2\pi)$. Equation~\eqref{eq:Eigenvalue_problem} represents the deterministic forward operator: given the damage condition described by $\mathbf{z}$, the modal properties $\{\mathbf{f}, \bm{\Phi}\}$ are uniquely determined by solving the generalized eigenvalue problem.

\subsection{The stochastic inverse problem: damage identification under uncertainty}
\label{sec:Inverse_Problem}
In SHM, time-domain vertical vibration signals $\ddot{\text{w}}(t)$ are acquired from a sparse network of accelerometers and processed using OMA techniques \cite{OMA_Feup2011}.
The output of this procedure forms an observation vector $\mathbf{m} = \text{flat}(\{\mathbf{f}, \bm{\Phi}\})\in \mathbb{R}^{d}$, where the symbol $\text{flat}(\cdot)$ indicates the flattening operation to produce a list of observed features with length $d = n_{m} \times (n_{s} +1)$, $\mathbf{f} \in \mathbb{R}^{n_{m}}$ contains the first $n_{m}$ natural frequencies of the system, and $\bm{\Phi} \in \mathbb{R}^{n_{m}\times n_{s}}$ is a matrix that contains $n_{m}$ $n_{s}$-dimensional mode shapes, measured at the $n_s$ sensors.
We work with the first $n_{m}$ eigenvectors since lower-frequency modes often exhibit higher accuracy and contain the most relevant information.
The dimension $n_{s}$ is the number of available sensors.
Under the linear elasticity assumption for early-stage damage, these intrinsic properties extracted via OMA are mathematically equivalent to the exact solution of the undamped generalized eigenvalue problem introduced in Section~\ref{sec:Parametrization}.

To infer the presence of damage, we map the measurements $\mathbf{m}$ back to the physical damage condition properties $\mathbf{z}$.
Due to aleatoric uncertainty (e.g., sensor noise, environmental variability) and limited spatial resolution ($n_s$), this inverse mapping is ill-posed and non-deterministic.
Thus, instead of seeking a singular point estimate, we formulate the problem probabilistically via Bayesian inference to infer the conditional posterior PDF $p(\mathbf{z}|\mathbf{m})$.

\section{The Sci-ML copula VAE}
\label{sec:methodology}
\subsection{Introduction}
\label{sec:method_intro}
\begin{figure*}[htbp]
	\centering
	\includegraphics[width=0.95\linewidth]{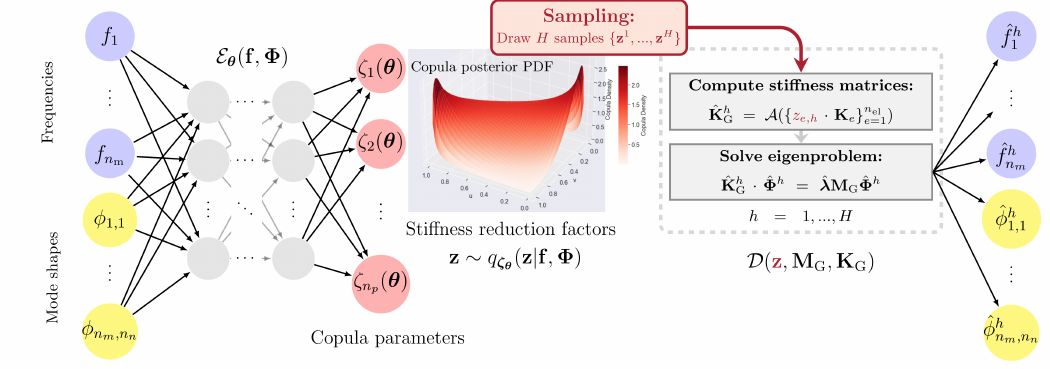}
	\caption{Illustrative scheme of the proposed VAE architecture. The encoder $\mathcal{E}_{\bm{\theta}}(\cdot)$ estimates the parameters of a distributional model describing the latent space (stiffness reduction factors, $\mathbf{z}$). The sampling layer builds the posterior PDF and draws samples that are fed to the decoder. The decoder $\mathcal{D}(\cdot)$ updates the global stiffness matrix and solves the generalized eigenvalue problem.}
	\label{fig:VAE_architecture}
\end{figure*}

To infer the conditional posterior distribution $p(\mathbf{z} | \mathbf{m})$ defined in Section~\ref{sec:Inverse_Problem}, we propose a variational autoencoder (VAE) architecture.
Figure~\ref{fig:VAE_architecture} shows a schematic illustration of the proposed architecture.
The following subsections describe the three main components of the VAE.

\subsection{The encoder}
\label{sec:encoder}
The encoder, denoted as $\mathcal{E}_{\bm{\theta}}$, is a fully connected neural network parameterized by weights and biases $\bm{\theta}$.
It maps the observed dynamic response—the flattened vector of natural frequencies and mode shapes $\mathbf{m} = [\mathbf{f}, \bm{\Phi}]$—into the probabilistic latent space of the stiffness reduction factors $\mathbf{z}$.
Rather than inferring a deterministic point estimate for the damage condition, the encoder maps the observations to the parameters $\bm{\zeta}(\bm{\theta})$ of a parametric joint posterior PDF, such that $\mathbf{z} \sim q_{\bm{\zeta}_{\bm{\theta}}}(\mathbf{z}|\mathbf{f}, \bm{\Phi})$.
Mathematically, the mapping is defined as:
\begin{equation}
	\mathcal{E}_{\bm{\theta}}: \mathbb{R}^{d} \rightarrow \bm{\zeta}(\bm{\theta}) \in \mathbb{R}^{n_{p}},
\end{equation}
where $\bm{\zeta}(\bm{\theta}) = \{\zeta^{1}(\bm{\theta}), ..., \zeta^{n_{p}}(\bm{\theta})\}$ contains the $n_{p}$ parameters required to describe the posterior PDF.

\subsection{The sampling layer}
\label{sec:sampling_layer}
The sampling layer connects the stochastic latent space and the deterministic physics-based decoder.
It constructs the posterior PDF from the parameters $\bm{\zeta}(\bm{\theta})$ estimated by the encoder and subsequently draws samples to feed the forward problem.

\textbf{Build the copula PDF:} we employ a Gaussian copula model to approximate the joint posterior density $q_{\bm{\zeta}_{\bm{\theta}}}(\mathbf{z}|\mathbf{f}, \bm{\Phi})$, capturing the dependence structure of the elements' damage state \cite{sklar1973random,trivedi2007copula}:
\begin{equation}
	q_{\bm{\zeta}_{\bm{\theta}}}(\mathbf{z}|\mathbf{f}, \bm{\Phi}) := c(F_1(z_1), \dots, F_{D}(z_{D})) \prod_{d=1}^{D} q_{d}(z_{d}|\mathbf{f}, \bm{\Phi}),
	\label{eq:joint_copula_density}
\end{equation}
where $c$ is the Gaussian copula density parameterized by a correlation matrix $\bm{\Sigma} = \mathbf{L L}^T$, defined through its Cholesky factor $\mathbf{L}$. $F_{d}(\cdot)$ is the cumulative distribution function (CDF) of the marginals $\{q_{d}\}_{d=1}^{D}$, which are assumed to be truncated Gaussian PDFs bounded by $\mathcal{S} = [z_{\text{lb}}, z_{\text{ub}}]$ to account for physical constraints. Each $z_d$ follows a Gaussian distribution with mean $\mu_d$ and standard deviation $\sigma_d$, truncated to $\mathcal{S} = [z_{\text{lb}},z_{\text{ub}}]$, with density and CDF:
\begin{equation}
	q_{d}(z)=\frac{\varphi\big((z-\mu_{d})/\sigma_{d}\big)}{\sigma_{d}\,Z_{d}},\qquad
	F_{d}(z)=\frac{\Psi\big((z-\mu_{d})/\sigma_{d}\big)-\Psi(\alpha_{d})}{Z_{d}},\qquad
	z\in\mathcal{S},
	\label{eq:marginal_tn}
\end{equation}
where $\varphi$ and $\Psi$ are the standard normal PDF and CDF, $\alpha_{d}=(z_{\text{lb}}-\mu_{d})/\sigma_{d}$, $\beta_{d}=(z_{\text{ub}}-\mu_{d})/\sigma_{d}$, and $Z_{d}=\Psi(\beta_{d})-\Psi(\alpha_{d})$. The parameters estimated by the encoder include the means and variances of these marginals, and the non-zero entries of $\mathbf{L}$, such that $\bm{\zeta}(\bm{\theta}) = \{ \bm{\mu}(\bm{\theta}), \bm{\sigma}(\bm{\theta}),\mathbf{l}(\bm{\theta})\}$.

\textbf{Draw samples from the copula PDF:}
\label{sec:sampling_copula}
we generate $H$ samples by first sampling a joint zero-mean Gaussian vector with covariance matrix $\mathbf{\Sigma}(\bm{\theta}) = \mathbf{L}(\bm{\theta}) \mathbf{L}(\bm{\theta})^{\top}$ by setting $\mathbf{y} =  \mathbf{L}(\bm{\theta})\mathbf{g}$, where $\mathbf{g}$ is a standard Gaussian vector ($\mathbf{g} \sim \mathcal{N}(\mathbf{0}, \mathbf{I})$).
We subsequently apply the copula transformation with the closed-form quantile function of each truncated Gaussian marginal:
\begin{equation}
	z_{d} = F_d^{-1}\big(\Psi(\text{y}_{d})\big) = \mu_{d}(\bm{\theta}) + \sigma_{d}(\bm{\theta})\,\Psi^{-1}\Big(\Psi(\alpha_{d}) + \Psi(\text{y}_{d})\,Z_{d}\Big), \qquad d=1,\dots,D,
	\label{eq:quantile}
\end{equation}
where $\mu_{d}(\bm{\theta})$ and $\sigma_{d}(\bm{\theta})$ are the estimated mean and standard deviation of variable $z_{d}$. This expression is differentiable with respect to the encoder outputs, and every sample lies within the support $\mathcal{S}$.
The obtained samples $\{\mathbf{z}^{1}, ..., \mathbf{z}^{H}\}$ represent $H$ stiffness reduction vectors drawn from the estimated posterior PDF $q_{\bm{\zeta}_{\bm{\theta}}}(\mathbf{z}|\mathbf{f}, \bm{\Phi})$.

\subsection{The physics-informed decoder}
\label{sec:decoder}
The decoder operator $\mathcal{D}(\cdot)$ receives the reduction factor samples $\{\mathbf{z}^{h}\}_{h=1}^{H}$ and scales the element stiffness matrices to assemble the global stiffness matrix $\hat{\mathbf{K}}^{h}_{\text{G}} = \hat{\mathbf{K}}_{\text{G}}(\mathbf{z}^{h})$ (Eq.~\eqref{eq:Global_Stiffness}). Next, we solve the generalized eigenvalue problem using the nominal mass matrix $\mathbf{M}_{\text{G}}$:
\begin{equation}
	\hat{\mathbf{K}}^{h}_{\text{G}} \hat{\bm{\Phi}}^{h}_{\text{full}} = \hat{\bm{\lambda}}^{h}_{\text{full}} \mathbf{M}_{\text{G}} \hat{\bm{\Phi}}^{h}_{\text{full}},
	\label{eq:full_eigenproblem}
\end{equation}
where $\hat{\bm{\Phi}}^{h}_{\text{full}}$ contains the eigenmodes, and $\hat{\bm{\lambda}}^{h}_{\text{full}} = (\hat{\bm{\omega}}^{h})^{2}$ contains the squared angular eigenfrequencies. This is transformed into a standard symmetric eigenvalue problem via Cholesky factorization of $\mathbf{M}_{\text{G}}$ \cite{strang2006linear}. The resulting eigenvectors are inherently mass-normalized, satisfying $(\hat{\bm{\Phi}}^{h}_{\text{full}})^{\top} \mathbf{M}_{\text{G}} \hat{\bm{\Phi}}^{h}_{\text{full}} = \mathbf{I}$, ensuring consistency with the experimental modes. Finally, we extract the first $n_{m}$ modes to match the input observations $\mathbf{m} = \{\mathbf{f}, \bm{\Phi}\}$.

\subsection{The loss function}
\label{sec:loss_description}
Training the VAE consists of minimizing a loss function based on the Evidence Lower BOund (ELBO) \cite{VInference_ELBO2017}, which balances data discrepancy (likelihood) and a regularization term that forces the approximate posterior to match the prior:
\begin{equation}
	\mathcal{L}_{\text{ELBO}}(\bm{\theta}) \approx \mathcal{L}_{\text{freq}}(\bm{\theta}) + \mathcal{L}_{\text{MAC}}(\bm{\theta}) + \gamma^2\mathcal{L}_{\text{PDF}}(\bm{\theta}).
	\label{eq:loss}
\end{equation}
The two data misfit terms (frequency and mode-shape) are scaled by the variance of the error of each measured quantity:
\begin{equation}
	\begin{aligned}
		\mathcal{L}_{\text{freq}}(\bm\theta)&=\frac{1}{N_{\text{train}}\cdot H}\sum_{i=1}^{N_{\text{train}}}\sum_{h=1}^{H}
		\sum_{j=1}^{n_m}\frac{\big(\log f_{i,j}-\log\hat f_{i,j}^{h}\big)^2}{2\,\sigma_{f,j}^2},\\
		\mathcal{L}_{\text{MAC}}(\bm\theta)&=\frac{1}{N_{\text{train}}\cdot H}\sum_{i,h,j}
		\frac{1-\mathrm{MAC}_j\big(\bm\Phi_i,\hat{\bm\Phi}_i^{h}\big)}{2\,\sigma_{\Phi,j}^2},
	\end{aligned}
	\label{eq:nll}
\end{equation}
where $\hat f_{i,j}^{h}$ and $\hat{\bm\Phi}_i^{h}$ are the modal data predicted by the decoder for the sample $\mathbf{z}^{h}$ of the $i$-th observation, i.e., the solution of Eq.~\eqref{eq:full_eigenproblem} with the damaged stiffness matrix $\hat{\mathbf{K}}_{\text{G}}(\mathbf{z}^{h})$ and the nominal mass matrix $\mathbf{M}_{\text{G}}$; the Modal Assurance Criterion (MAC) \cite{MAC} compares the measured and estimated mode shapes; and $\sigma_{f,j}^2$ and $\sigma_{\Phi,j}^2$ are the variances of the errors of the $j$-th logarithmic frequency and of the $j$-th mode shape.

The posterior PDF term $\mathcal{L}_{\text{PDF}}(\bm{\theta})$ measures the probability that a certain sample $\mathbf{z}^{h}$ belongs to the approximate posterior:
\begin{equation}
	\begin{aligned}
		\mathcal{L}_{\text{PDF}}(\bm{\theta})
		&= \frac{1}{N_{\text{train}}\cdot H}  \sum_{i=1}^{N_{\text{train}}}\sum_{h=1}^{H}\log q_{\bm{\zeta}_{\bm{\theta}}}( \mathbf{z}^{h}_{i}(\bm{\theta})|\mathbf{f}_{i},\bm{\Phi}_{i}).
	\end{aligned}
\end{equation}
This term acts as a regularizer applied to the latent space (stiffness reduction factors) by forcing the approximate posterior to match the prior distribution.

The optimal encoder parameters $\bm{\theta}^{*}$ are obtained by minimizing this total loss.

\section{Implementation}
\label{sec:implementation}
\subsection{Case study and finite element discretizations}
\label{sec:case_study}
The case study represents a reinforced concrete slab bridge, following previous studies on short-span bridges~\cite{moliner2017dynamic,luo2025joint}.
The deck has a span of $L=10$~m, a width of $B=11.6$~m and a depth of $h=0.77$~m, an elastic modulus $E=34.85$~GPa, and a mass per unit length of 31,301~kg/m.
Each bearing is represented by a vertical spring with nominal stiffness $K_{v}=9.5\times10^{6}$~kN/m, which corresponds to the intermediate support flexibility analyzed by~\cite{moliner2017dynamic} and lies within the range of commercially available elastomeric bearings ($10^{5}$--$10^{7}$~kN/m)~\cite{erduran2022effect}.
Table~\ref{tab:specs} summarizes the specifications of the structural system.
\begin{table}[htbp]
	\centering
	\caption{Nominal values of the case study.}
	\label{tab:specs}
	\begin{tabular}{ll}
		\toprule
		Item & Value \\
		\midrule
		Deck & Reinforced concrete slab, $L=10$~m, $B=11.6$~m, $h=0.77$~m \\
		Concrete & Elastic modulus $E=34.85$~GPa \\
		Mass & 31,301~kg/m \\
		Bending stiffness & $EI=1.54\times10^{10}$~N\,m$^2$, with $I=Bh^3/12$ \\
		Bearings & Vertical springs, $K_{v}=9.5\times10^{6}$~kN/m each \\
		\bottomrule
	\end{tabular}
\end{table}

\begin{figure}[htbp]
	\centering
	\includegraphics[width=\linewidth]{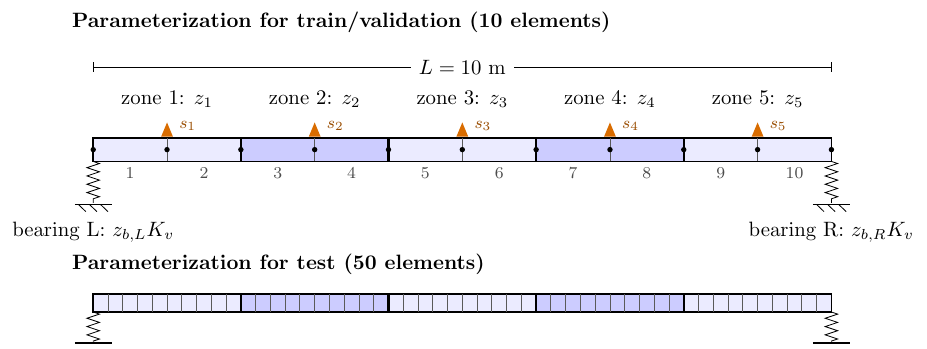}
	\caption{FE parameterization of the short-span bridge on vertical elastic bearings, with five deck zones and two bearings (stiffness factors $z_1,\dots,z_5$, $z_{b,L}$ and $z_{b,R}$). Top: 10-element model used for the train/validation datasets; bottom: 50-element model used for the test dataset. Five accelerometers ($s_1,\dots,s_5$) are placed at the centers of the zones.}
	\label{fig:beam_elements}
\end{figure}

We employ two FE discretizations of the bridge (Fig.~\ref{fig:beam_elements}):
\begin{itemize}
	\item \emph{10-element model} ($L_e=1$~m): Used to generate the train and validation datasets.
	\item \emph{50-element model} ($L_e=0.2$~m): Used to generate the test datasets.
\end{itemize}
Both models have the same geometry, material properties and bearings, and every node of the 10-element model is also a node of the 50-element model.
The deck is divided into $n_z=5$ damage zones of 2~m (shaded regions in Fig.~\ref{fig:beam_elements}), which comprise two elements of the 10-element model and ten elements of the 50-element model, so that each stiffness factor has the same physical meaning in both discretizations.
Five vertical accelerometers ($n_s=5$; $s_1,\dots,s_5$ in Fig.~\ref{fig:beam_elements}) are placed at the centers of the zones ($x=1$, 3, 5, 7 and 9~m), which are nodes of both models; hence, both models provide the mode shapes at the same locations.
We identify the first $n_m=5$ modes. The lower bound of the stiffness factors is $z_{\text{lb}}=0.45$, i.e., a maximum loss of stiffness of 55\%.

\subsection{Design of experiments}
A component (deck zone or bearing) is considered damaged when its stiffness factor is below the health limit $\eta=0.95$.
Every scenario belongs to one of the categories of Table~\ref{tab:doe}, defined by its number of damaged components, with the largest shares assigned to healthy scenarios and to damage at a single location.
Within each category, every component and every combination of components is damaged in the same number of scenarios, so that no location is favored; the factors of the damaged components span $[z_{\text{lb}},\eta)$, and those of the undamaged components lie in $[\eta,1]$, which represents the healthy state with small as-built variations.

\begin{table}[htbp]
	\centering
	\caption{Categories of the design of experiments: number of simultaneously damaged components, share of the training, validation and test datasets, and number of test scenarios.}
	\label{tab:doe}
	\begin{tabular}{lccc}
		\toprule
		Category & Damaged components & Share & Test scenarios \\
		\midrule
		Healthy & 0 & 30\% & 358 \\
		Single damage & 1 & 30\% & 362 \\
		Double damage & 2 & 15\% & 179 \\
		Triple damage & 3 & 10\% & 120 \\
		Multiple damage & 4 or more & 5\% & 61 \\
		Uniform & any, $\mathbf{z}\sim\mathcal{U}[z_{\text{lb}},1]^D$ & 10\% & 120 \\
		\bottomrule
	\end{tabular}
\end{table}

The training ($N_{\text{train}}=14{,}000$) and validation (2,400) data are generated with a 10-element model of the bridge (two elements per zone). The test dataset (1,200 scenarios) follows the same design of experiments, but it is generated with a finer 50-element model (ten elements per zone, see Fig.~\ref{fig:beam_elements}), which introduces an additional discretization error. In every dataset, the same factor $z_d$ multiplies the bending stiffness of all the elements of zone $d$, so that the damage is spread over the entire zone.

To analyze the behavior of the method under distribution shift, we generate an additional test dataset (test dataset B) with the same 1,200 scenarios and the same 50-element model, but in every damaged zone the loss of stiffness is concentrated in a subregion of length $\ell_d$, at a random position within the zone, while the rest of the zone keeps its nominal stiffness. The stiffness factor $s$ of the subregion is chosen so that the zone keeps the same overall stiffness factor $z_d$: if a fraction $\alpha=\ell_d/L_{d}$ of the zone has stiffness factor $s$, both zones are equally flexible when \cite{sinha2002simplified}
\begin{equation}
	\frac{1}{z_d}=\frac{\alpha}{s}+(1-\alpha)
	\qquad\Longrightarrow\qquad
	s=\frac{\alpha\,z_d}{1-(1-\alpha)\,z_d}.
	\label{eq:localized}
\end{equation}
For example, a zone with $z_d=0.6$ can be damaged over 0.4~m with $s=0.23$, or over 1.0~m with $s=0.43$.
For each state, we solve the generalized eigenvalue problem to extract the first $n_m = 5$ mass-normalized mode shapes and natural frequencies.

\subsection{Modeling error and measurement noise}
\paragraph{Modeling error.} The mass of each element of the structures used to generate the data is perturbed as:
\begin{equation}
	m_e^{\text{true}} = m_e^{\text{nom}} \left(1 + \epsilon\, \xi_e\right), \quad e = 1, \dots, n_{\text{el}},
	\label{eq:modeling_error}
\end{equation}
where $m_e^{\text{nom}}$ is the nominal mass of the $e$-th element, $\xi_e$ is an independent sample of a standard Gaussian random variable, and $\epsilon=0.05$ is the level of the modeling error (5\%), with a new realization for every simulated structure. The perturbed (true) masses are used to generate the datasets, whereas the physics-informed VAE only knows the nominal values.

\paragraph{Measurement noise.} To account for measurement noise and OMA imprecision, we apply multiplicative zero-mean Gaussian noise ($\sigma_f = 0.025$) to the frequencies \cite{xia2002damage} and additive noise ($\sigma_\phi = 0.05$) to the peak amplitude of each mode shape \cite{reynders2008uncertainty}.

\subsection{Architecture design and training specifications}
\label{Sec:NN_specifications}
We implement the Sci-ML copula VAE using TensorFlow \cite{abadi2016tensorflow}. The probabilistic encoder $\mathcal{E}_{\bm{\theta}}$ is a fully connected network with three hidden layers (128, 256 and 128 neurons), using rectified linear unit (ReLU) activation and He Uniform initialization \cite{hanin2018start}, with 75,306 trainable parameters.
The output layer splits into branches that estimate the means and standard deviations of the marginals and the entries of $\mathbf{L}$.
The samples are drawn with the closed-form quantile of Eq.~\eqref{eq:quantile}.
We use batches of 256 observations and $H=6$, i.e., 1,536 samples per gradient step.
The physics-based decoder uses vectorized tensor operations to assemble the global stiffness matrix and a differentiable symmetric eigenvalue solver to compute the modal properties.
We train using the Adam optimizer~\cite{Adam2015} with a learning rate of $5\times10^{-4}$.
The weight of the regularization term is fixed to $\gamma=1$, for which Eq.~\eqref{eq:loss} is the negative ELBO (Appendix~\ref{app:gamma} shows the effect of other values).
The training stops after 2,790 epochs. We keep the weights with the lowest validation negative ELBO.
The training took about one hour (2,790 epochs of about 1.4~s) on a server with one NVIDIA L40S graphics processing unit (GPU) and two AMD EPYC 7763 central processing units (CPUs); the encoder runs on the GPU, and the eigenvalue problems (1,536 per gradient step) are solved on the CPU in double precision. Once trained, the posterior of a new observation is obtained with a single forward pass of the encoder in less than 1~ms.

\begin{figure*}[htbp]
	\centering
	\includegraphics[width=\linewidth]{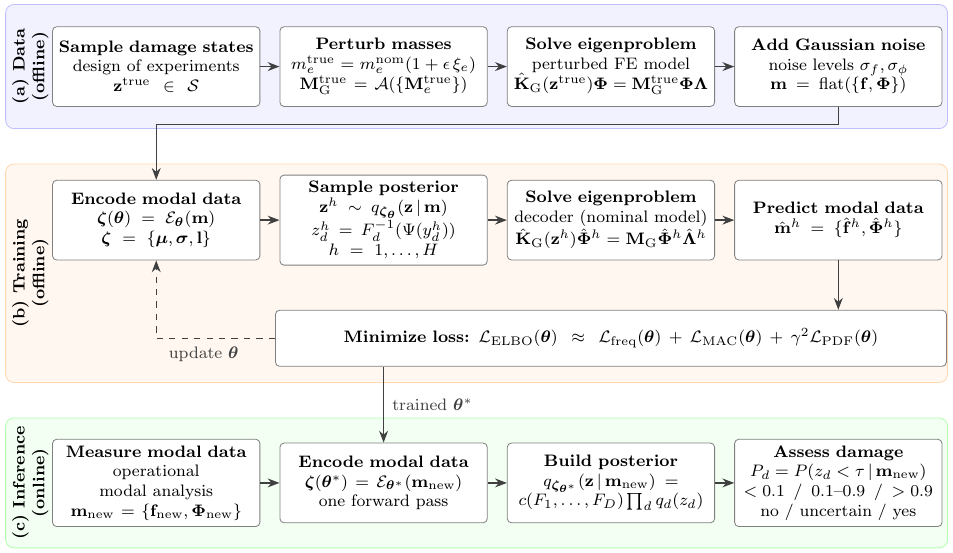}
	\caption{Methodology: (a) generation of the synthetic data, (b) physics-informed training of the encoder through the embedded eigenvalue solver, without the true damage states, and (c) online damage identification with the trained encoder.}
	\label{fig:flowchart}
\end{figure*}

After the training, the model is ready for the online inference task. Figure~\ref{fig:flowchart} summarizes the whole methodology: the generation of the synthetic data and the physics-informed training of the encoder (offline), and the damage identification or inference phase (online).

\section{Results and discussion}
\label{sec:results}
We evaluate the proposed VAE model on the two test datasets, which take no part in the training (see Section ``Design of experiments'').
To analyze the performance, we calculate point-estimate accuracy using the mean squared error (MSE) and mean absolute error (MAE), probabilistic calibration using the 95\% coverage probability and mean absolute calibration error (MACE), and distributional quality using the Average log-likelihood (ALL) \cite{bektacs2010comparison}, defined in Appendix~\ref{sec:metrics}.

\subsection{Performance analysis}
\label{sec:gamma_comparison}
Table~\ref{tab:metrics} reports the metrics of Eqs.~\eqref{eq:mse_mae}--\eqref{eq:mace} on the validation dataset and on both test datasets.
On the original test dataset, the metrics are close to those on the validation dataset, which is generated with the same FE model as the training data; hence, the discretization error of the 50-element model has a small effect.
The localized damage of test dataset B, which is not represented in the training data, slightly increases the errors (MSE of 0.00646 against 0.00570) and reduces the 95\% coverage from 0.900 to 0.885.
In both test datasets, 89--90\% of the true stiffness factors lie within their 95\% credible intervals, i.e., the posteriors are slightly overconfident, as is typical of variational inference.

\begin{table}[htbp]
	\centering
	\caption{Performance metrics on the validation dataset (2,400 observations) and on the test datasets (1,200 observations each).}
	\label{tab:metrics}
	\begin{tabular}{lccc}
		\toprule
		Metric & Validation dataset & Test dataset & Test dataset B \\
		\midrule
		MSE & 0.00584 & 0.00570 & 0.00646 \\
		MAE & 0.0607 & 0.0601 & 0.0616 \\
		ALL & 1.715 & 1.723 & 1.661 \\
		95\% coverage, $C(0.95)$ & 0.895 & 0.900 & 0.885 \\
		MACE & 0.107 & 0.101 & 0.103 \\
		\bottomrule
	\end{tabular}
\end{table}

\subsection{Damage detection and false positives}
\label{sec:detection}
Rather than a binary decision based on a point estimate, we exploit the posterior to quantify the evidence of damage~\cite{vanik2000bayesian}.
For each component, we compute the posterior probability of a significant damage:
\begin{equation}
	P_d=P\big(z_d<\tau\,|\,\mathbf{m}\big)=F_d(\tau),
	\label{eq:pd}
\end{equation}
where $\tau$ is the limit of significant damage, and $P_d$ is obtained directly from the marginal posterior (Eq.~\eqref{eq:marginal_tn}).
We set $\tau=0.80$ according to the sensitivity of the measurements: a loss of stiffness of $1-\tau$ in a single component changes the first five natural frequencies by about as much as the standard deviation of their errors (measurement noise and modeling error), so that smaller losses cannot be reliably identified.

Based on $P_d$, each component is classified as \emph{no significant damage} ($P_d<P_{\ell}$), \emph{significant damage} ($P_d>P_{u}$) or \emph{uncertain} ($P_{\ell}\le P_d\le P_{u}$), with $P_{\ell}=0.1$ and $P_{u}=1-P_{\ell}$, i.e., the same confidence level for both verdicts, fixed a priori as for the credible intervals.
These limits are meaningful because $P_d$ is a calibrated probability: on the validation dataset, the fraction of components with $z_d^{\text{true}}<\tau$ matches $P_d$.
An uncertain verdict indicates that the measurements do not contain enough information to decide, so that the component should be inspected or monitored with additional data.
Other limits only change the share of uncertain verdicts; the optimal limits for a given structure depend on the consequences of each wrong decision (e.g., the cost of an unnecessary inspection versus that of an undetected damage), which can be included through decision analysis~\cite{kamariotis2022value}.
Accordingly, a false positive (FP) is a component without significant damage ($z_d^{\text{true}}\ge \tau$) classified as significant damage, and a false negative (FN) is a component with significant damage ($z_d^{\text{true}}<\tau$) classified as no significant damage.

\begin{table}[htbp]
	\centering
	\caption{Verdicts on the components of the test dataset, based on the posterior probability of significant damage, $P_d$ (Eq.~\eqref{eq:pd}). Each row adds up to 100\%; wrong confident verdicts in bold.}
	\label{tab:verdicts}
	\small\setlength{\tabcolsep}{4pt}
	\begin{tabular}{llcccc}
		\toprule
		True state & $z_d^{\text{true}}$ & Components & No significant & Uncertain & Significant \\
		& & & damage ($P_d<P_{\ell}$) & & damage ($P_d>P_{u}$) \\
		\midrule
		Healthy & $\ge\eta$ & 6,281 & 74.7\% & 25.3\% & \textbf{0.0\%} \\
		Light damage & 0.90--0.95 & 226 & 56.2\% & 42.9\% & \textbf{0.9\%} \\
		& 0.85--0.90 & 203 & 33.5\% & 66.0\% & \textbf{0.5\%} \\
		& 0.80--0.85 & 197 & 21.8\% & 74.6\% & \textbf{3.6\%} \\
		Significant damage & 0.70--0.80 & 429 & \textbf{4.9\%} & 76.0\% & 19.1\% \\
		& $<0.70$ & 1,064 & \textbf{0.2\%} & 23.3\% & 76.5\% \\
		\bottomrule
	\end{tabular}
\end{table}

Table~\ref{tab:verdicts} shows that the confident verdicts are almost always correct: only 0.2\% of the components without significant damage are classified as significant damage (FP), and 1.5\% of the components with significant damage are classified as no significant damage (FN); test dataset B gives similar values (0.5\% and 2.5\%).
The uncertain verdicts concentrate where the measurements are less informative: light damage ($\tau\le z_d^{\text{true}}<\eta$), whose effect on the modal data is below the error level; damage just below $\tau$; and the end zones of the deck, where the bending moment is small in every mode.
This is a limitation of the information contained in the measurements (five sensors and five modes), not of the method: the posterior reports when the data cannot distinguish a damaged component from a healthy one, instead of providing a confident but wrong decision (see also Appendix~\ref{sec:modes}).

\begin{figure*}[htbp]
	\centering
	\includegraphics[width=\linewidth]{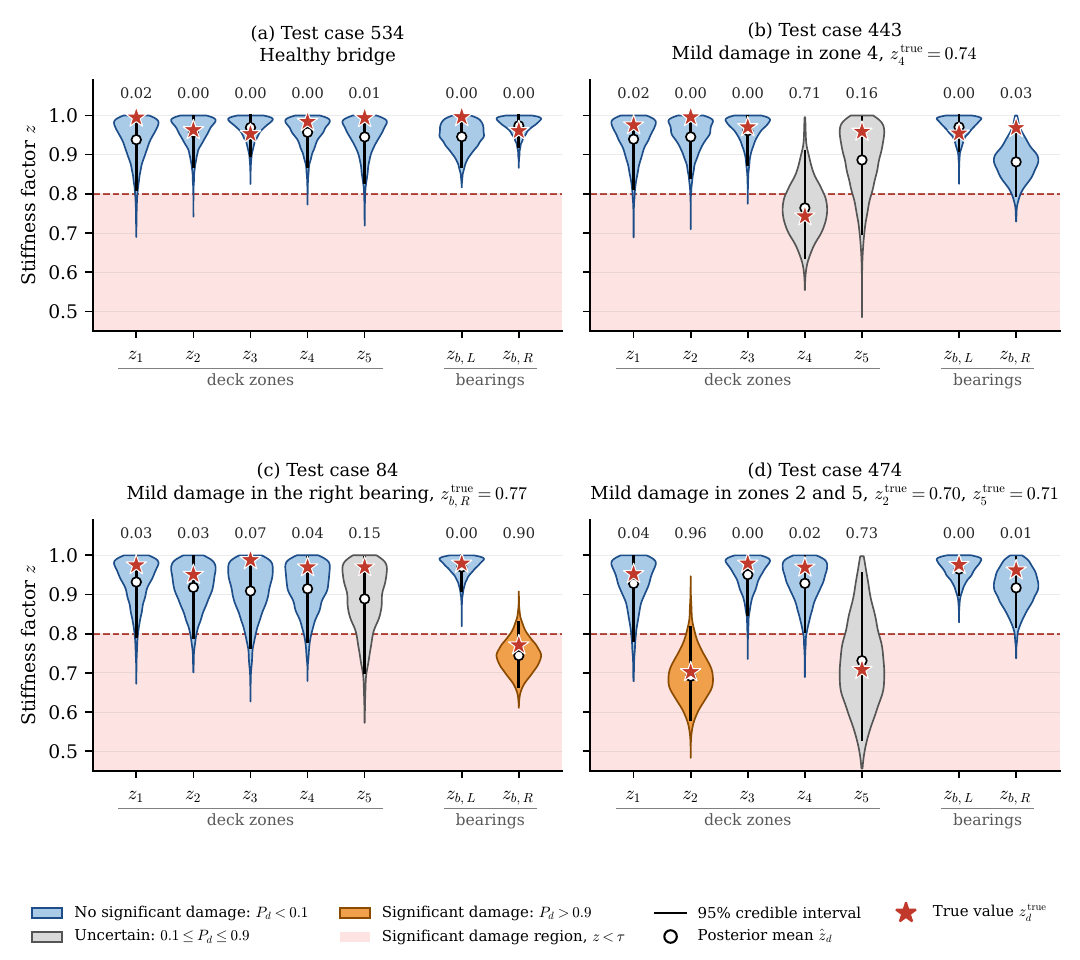}
	\caption{Posteriors of the stiffness factors (violin plots) of four scenarios of test dataset B, colored according to the verdict: no significant damage ($P_d<P_{\ell}$), uncertain, or significant damage ($P_d>P_{u}$). Numbers on top: $P_d=P(z_d<\tau\,|\,\mathbf{m})$. (a) Healthy bridge; (b) mild damage in zone 4; (c) mild damage in the right bearing; (d) mild damage in zones 2 and 5.}
	\label{fig:cases_well}
\end{figure*}

\begin{figure*}[htbp]
	\centering
	\includegraphics[width=\linewidth]{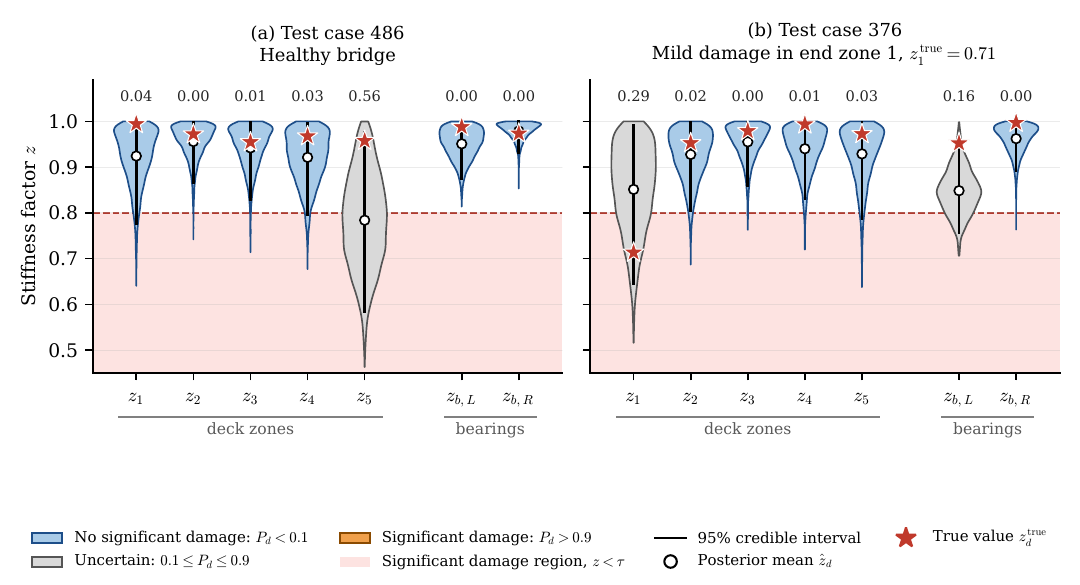}
	\caption{Posteriors of the stiffness factors (violin plots) of two scenarios of test dataset B in which a decision based only on the posterior mean would be wrong: (a) healthy end zone 5; (b) damage in end zone 1, localized near the left support. Colors and numbers as in Fig.~\ref{fig:cases_well}.}
	\label{fig:cases_fail}
\end{figure*}

Figs.~\ref{fig:cases_well} and~\ref{fig:cases_fail} show the marginal posteriors (violin plots) of six scenarios of test dataset B, in which the damage is localized within each zone (distribution shift). Each violin represents the posterior of one component, colored according to its verdict, with its 95\% credible interval, its mean $\hat z_d$, the true value $z_d^{\text{true}}$ and the value of $P_d$ on top; the shaded region corresponds to significant damage ($z<\tau$).
Fig.~\ref{fig:cases_well} shows (a) a healthy bridge, in which no component has significant damage; (b) a mild damage in zone 4, close to $\tau$, whose posterior is centered on the true value but with an uncertain verdict; (c) a mild damage in the right bearing, classified as significant damage; and (d) a mild damage in zones 2 and 5, where zone 2 is classified as significant damage and the end zone 5 is uncertain, with a much wider posterior.

Fig.~\ref{fig:cases_fail} shows two scenarios in which a decision based only on the posterior mean would be wrong, both in the end zones of the deck. In (a), the posterior mean of the healthy zone 5 is below $\tau$, but its posterior is wide and the verdict is uncertain, not a false positive. In (b), the loss of stiffness of zone 1 is concentrated near the left support, and its posterior mean is above $\tau$, but the verdict is uncertain, not a false negative; the adjacent bearing is also uncertain, since a loss of stiffness of an end zone and a softer bearing have similar effects on the modal data.

\subsection{Analysis of the estimated posteriors for specific test scenarios}
\label{subsec:rqmc_comparison}
To validate the quality of the learned posterior, we compare the VAE estimates with a reference posterior computed with sequential Monte Carlo (SMC) sampling \cite{ching2007transitional}, with 4,000 particles and the same likelihood as in the training (Eq.~\eqref{eq:nll}).

\begin{figure*}[htbp]
	\centering
	\begin{subfigure}[b]{0.49\textwidth}
		\centering
		\includegraphics[width=\linewidth]{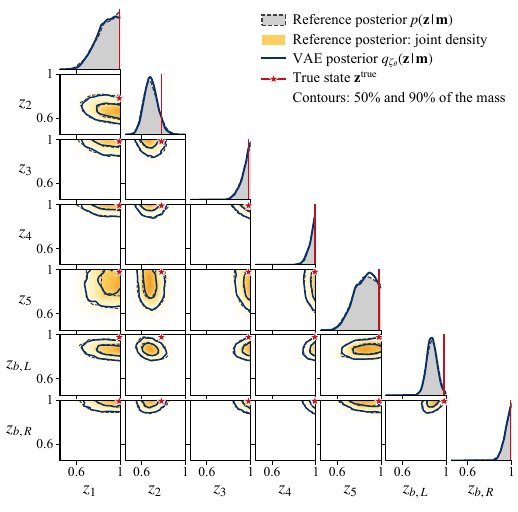}
		\caption{Test dataset, mild damage: $z_2^{\text{true}}=0.78$}
		\label{fig:overlay_P3}
	\end{subfigure}
	\hfill
	\begin{subfigure}[b]{0.49\textwidth}
		\centering
		\includegraphics[width=\linewidth]{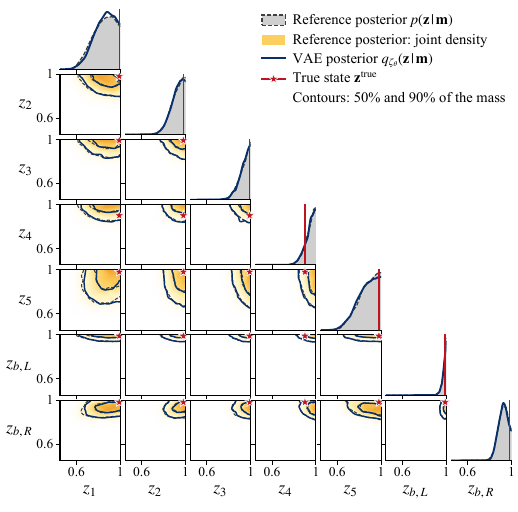}
		\caption{Test dataset, light damage: $z_4^{\text{true}}=0.90$}
		\label{fig:overlay_P10}
	\end{subfigure}
	\\[2ex]
	\begin{subfigure}[b]{0.49\textwidth}
		\centering
		\includegraphics[width=\linewidth]{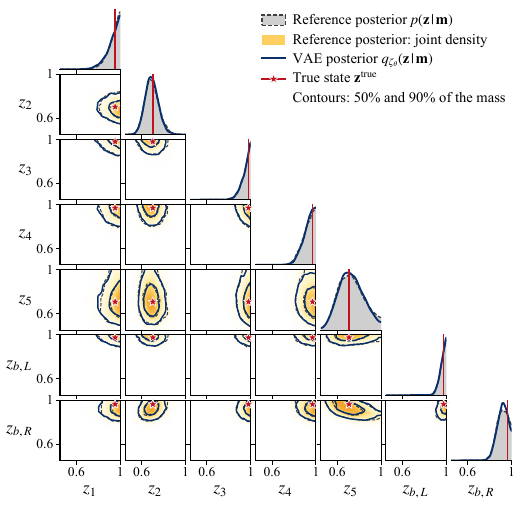}
		\caption{Test dataset B, mild damage in zones 2 and 5}
		\label{fig:overlay_474}
	\end{subfigure}
	\hfill
	\begin{subfigure}[b]{0.49\textwidth}
		\centering
		\includegraphics[width=\linewidth]{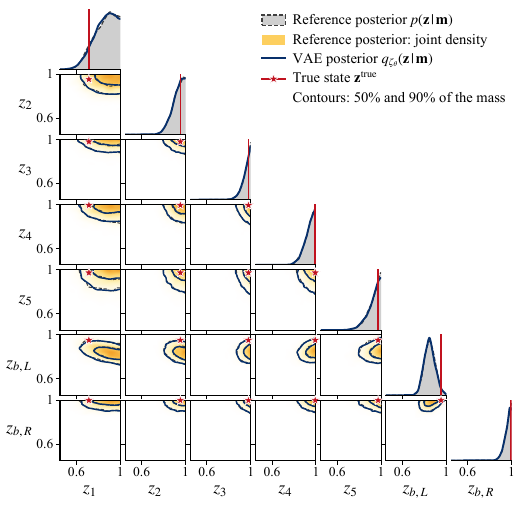}
		\caption{Test dataset B, mild damage in end zone 1}
		\label{fig:overlay_376}
	\end{subfigure}
	\caption{Joint posteriors of four test scenarios: VAE posterior $q_{\bm{\zeta}_{\bm{\theta}^{*}}}(\mathbf{z}|\mathbf{m}_{t})$ (solid lines) and reference posterior $p(\mathbf{z}|\mathbf{m}_{t})$ (gray marginals, colored joint densities and dashed contours). Red: true damage state $\mathbf{z}^{\text{true}}$.}
	\label{fig:posterior_comparison}
\end{figure*}

Once we have defined the reference posterior $p(\mathbf{z}|\mathbf{m})$, we can compare it with the estimated posterior $q_{\bm{\zeta}_{\bm{\theta}}}(\mathbf{z}|\mathbf{m})$.
Fig.~\ref{fig:posterior_comparison} compares both posteriors for four scenarios: two scenarios of the test dataset with mild and light damage in a single zone ($z_2^{\text{true}}=0.78$ and $z_4^{\text{true}}=0.90$), and the two scenarios of test dataset B of Figs.~\ref{fig:cases_well}(d) and~\ref{fig:cases_fail}(b), i.e., the mild damage in zones 2 and 5, correctly identified, and the mild damage of the end zone 1, with an uncertain verdict.
Each subfigure includes the marginal distribution of each reduction factor $z_d$, $d=1,\dots,D$, (diagonal positions), and the 50\% and 90\% contours of the joint distribution of every variable pair (off-diagonal positions).
In every two-dimensional PDF, we mark the exact values of the stiffness factors used to produce the test scenario with a red star.

The figures demonstrate the flexibility of the copula-based posterior to capture the topology of the true distribution.
In all the scenarios, the VAE posterior closely matches the reference posterior (mean Wasserstein-1 distance between the marginals below 0.004 in the four scenarios of Fig.~\ref{fig:posterior_comparison} and in those of Figs.~\ref{fig:cases_well} and~\ref{fig:cases_fail}).
The mild damage of Fig.~\ref{fig:overlay_P3} is detected and quantified. The light damage of Fig.~\ref{fig:overlay_P10} ($z_4^{\text{true}}=0.90$) is at the detection limit of the measurements: both posteriors keep most of their probability near the healthy state, while the true value remains inside the credible interval. The VAE thus reproduces the uncertainty of the exact Bayesian solution instead of providing an overconfident estimate.
The existing cross-correlations between pairs of variables (see e.g., $z_{5}$ and $z_{b,R}$ in Fig.~\ref{fig:overlay_474}, and $z_{1}$ and $z_{b,L}$ in Fig.~\ref{fig:overlay_376}) are successfully identified by the Gaussian copula $q_{\bm{\zeta}_{\bm{\theta}^{*}}}(\mathbf{z}|\mathbf{m}_{t})$.
In Fig.~\ref{fig:overlay_376}, the damage of the end zone and the softening of the adjacent bearing have similar effects on the modal data, so that both factors partly compensate each other. The reference posterior is equally wide, which shows that the uncertain verdict is caused by the limited information of the measurements, not by the approximation of the VAE.

\subsection{Comparison with a surrogate-based decoder}
\label{sec:surrogate}
The proposed method builds on the Gaussian mixture autoencoder of~\cite{Navamuel_wes2025} and the Gaussian copula VAE of~\cite{Navamuel2026copula2d}, in which the forward operator is approximated by a surrogate neural network (decoder), pre-trained with labeled data and then frozen while the encoder is trained. To compare both strategies, we replace the eigenvalue solver by a surrogate neural network, pre-trained with the same training and validation data, while the rest of the architecture remains unchanged. Table~\ref{tab:surrogate} compares both approaches for the training dataset of the case study (14,000 simulations) and for a reduced dataset of 500 simulations drawn at random from it.

\begin{table*}[htbp]
	\centering
	\caption{Comparison of the surrogate-based decoder and the proposed method on the test dataset. $\Delta$: relative change of each metric when the surrogate is replaced by the eigenvalue solver.}
	\label{tab:surrogate}
	\small\setlength{\tabcolsep}{4pt}
	\begin{tabular}{llcccccc}
		\toprule
		Simulations & Decoder & Parameters & MSE & MAE & ALL & $C(0.95)$ & MACE \\
		\midrule
		14,000 & Surrogate & 88,946 & 0.00572 & 0.0601 & 1.725 & 0.902 & 0.101 \\
		       & Eigenvalue solver & 75,306 & 0.00570 & 0.0601 & 1.723 & 0.900 & 0.101 \\
		       & $\Delta$ & & $-0.3\%$ & $-0.1\%$ & $-0.1\%$ & $-0.3\%$ & $+0.6\%$ \\
		\midrule
		500    & Surrogate & 88,946 & 0.00781 & 0.0723 & 1.393 & 0.834 & 0.185 \\
		       & Eigenvalue solver & 75,306 & 0.00668 & 0.0642 & 1.636 & 0.882 & 0.118 \\
		       & $\Delta$ & & $-14.5\%$ & $-11.2\%$ & $+17.5\%$ & $+5.7\%$ & $-36.0\%$ \\
		\bottomrule
	\end{tabular}
\end{table*}

With 14,000 simulations, both approaches give comparable metrics, i.e., the surrogate is accurate enough when it is trained with a large dataset covering the entire parameter space. However, with 500 simulations, the proposed method outperforms the surrogate approach in every metric: it reduces the error metrics (MSE by 14.5\%, MAE by 11.2\% and MACE by 36.0\%) and yields a higher ALL (+17.5\%) and 95\% coverage (from 0.834 to 0.882). The credible intervals of both approaches have practically the same mean width, but those of the surrogate approach contain the true values less often: the approximation error of the surrogate displaces its posterior, and since this error is not included in the likelihood, the posterior becomes overconfident. Hence, the exact eigenvalue solver improves the reliability of the uncertainty quantification and the point estimates, mainly when data are scarce.

An existing limitation of the method is its inability to capture strongly multimodal marginals. The expressiveness limitations of using single Gaussian marginal distributions prevent from satisfactorily capturing the true posterior modes. Future work requires exploring more expressive continuous PDFs to capture marginal multimodality, such as mixtures of bounded distributions \cite{wasserman2024stabilizing}.

\section{Conclusions and future work}
\label{sec:conclusions}
\textbf{Conclusions:} This study has introduced a novel Scientific Machine Learning (SciML) framework that synthesizes variational Bayesian inference with differentiable structural dynamics to address the long-standing challenge of uncertainty quantification in vibration-based damage identification. By reformulating the structural health monitoring (SHM) inverse problem as a generative task constrained by physical laws, we have demonstrated that robust damage assessment is possible even in the presence of significant aleatoric noise and sparse sensing.

The integration of a differentiable numerical eigenvalue solver as the decoder poses an improvement over standard black-box autoencoders. By solving the generalized eigenvalue problem directly within the computational graph, the model ensures physical consistency without requiring a surrogate training phase. This approach reduces the number of trainable parameters, constraining the search space and thus lightening the training process and generalization capabilities.

The adoption of a Gaussian copula to model the latent-space posterior distribution provides an efficient parameterization of the dependencies among the stiffness-reduction factors of the different structural elements. This capability is crucial for identifying localized damage in interconnected systems where stiffness reductions are rarely independent.
Unlike Gaussian mixture models, which suffer from exponential parameter growth in higher dimensions, the copula model maintains a fixed complexity as the finite-element discretization increases.
This capability makes our proposed method scalable to large-scale infrastructures where high-dimensional latent spaces are required to describe distributed damage patterns.

We validated the method on a synthetic short-span bridge on elastic bearings, with seven stiffness factors (five deck zones and two bearings), including measurement noise and modeling errors, and two test datasets generated with a finer FE discretization, to demonstrate the robustness of the method.
The methodology can be generalized as it only requires that the stiffness matrix is a weighted sum of matrices linked to the unknown parameters (Eq.~\eqref{eq:Global_Stiffness}).
Hence, it can be applied to other structures and damage parameterizations (e.g., continuous bridges with intermediate bearings, slender structures such as wind turbines or truss frames) by extending the latent variables.

The results revealed that providing a posterior distribution of the target features yields a measure of the inherent uncertainty, preventing overconfident diagnostics typical in deterministic methods. The alignment between the estimated credible intervals and the ground truth suggests that the Gaussian copula can reliably enclose the true structural state even in presence of modeling and measurement corruption errors.

\textbf{Limitations and future research:} Despite the advantages of our proposed method, there exist several research lines for improvement. First, the validation is purely numerical: although we account for modeling and measurement error, the method has not been validated with laboratory or field measurements, and the environmental variability has not been included in the present study.
Second, the methodology is restricted to systems governed by linear structural behavior, which is in general the case in the context of early damage identification in short-span bridges.
Handling nonlinearities requires re-formulating the forward problem and the use of substructure coupling techniques~\cite{jensen2019sub}.
Third, the eigenvalue problem is solved at every training step; for large 2D or 3D FE models, sparse eigensolvers that compute only the first modes, reduced-order models, or a surrogate of the forward operator would be required to keep the training cost affordable. Finally, a limitation of the Gaussian-marginal based distributional model is its inability to capture strongly multimodal marginal posteriors.
Future work requires exploring more complex continuous PDFs that capture marginal multimodality, such as mixtures of known unimodal distributions.
This is a challenging task, primarily due to the need to constrain unbounded distributions and the expensive computation of the inverse CDF~\cite{Navamuel_wes2025}.
These steps are required to generate samples during the training phase.
We consider exploring mixtures of the Beta or Kumaraswamy families \cite{wasserman2024stabilizing}, while keeping the dependence structure model via the copula.
These PDFs have the advantage of being bounded and supporting the interval $[0,1]$, and of having relatively tractable CDFs compared to Gaussian mixtures.
We also consider, as an alternative, the use of normalizing flows to enhance the topological flexibility of the approximate posterior.

Furthermore, the transition from synthetic validation to real-world application remains a critical step.
The proposed work provides an uncertainty-aware approach that efficiently handles the inherent uncertainty of the inverse problem, and it employs modal properties, which are proven to be robust to environmental and operational variability.
While experimental validation is the ultimate goal, the use of a high-fidelity synthetic dataset allows for the calculation of probabilistic metrics against a known true posterior, which is unfeasible with experimental data.
However, subsequent studies will focus on incorporating environmental and operational variability as inputs into the VAE architecture and on validating the model against experimental data from an experimental benchmark and from operating bridges.
Integrating the proposed system (stiffness reduction factors) identification strategy within an input load estimation scheme trained with short-term acceleration signals is scoped as an alternative to identify the affecting environmental and operational conditions.

\appendix
\section{Performance metrics}
\label{sec:metrics}
For each observation $\mathbf{m}_t$ ($t=1,\dots,N_{\text{test}}$) of the test dataset, with true damage condition $\mathbf{z}^{\text{true}}_t$, the trained encoder provides the marginal posteriors $q_{t,d}$ (Eq.~\eqref{eq:marginal_tn}), with CDF $F_{t,d}$, mean $\mu_{t,d}$ and standard deviation $\sigma_{t,d}$ ($d=1,\dots,D$). The metrics are averaged over the test dataset:
\begin{itemize}
	\item MSE and MAE of the point estimate $\hat z_{t,d}$, the mean of the marginal posterior, which has a closed form:
	\begin{equation*}
		\hat z_{t,d}=\int_{z_{\text{lb}}}^{z_{\text{ub}}} z\,q_{t,d}(z)\,\mathrm{d}z
		=\mu_{t,d}+\sigma_{t,d}\,\frac{\varphi(\alpha_{t,d})-\varphi(\beta_{t,d})}{Z_{t,d}},
	\end{equation*}
	\begin{equation}
		\mathrm{MSE}=\frac{1}{N_{\text{test}}D}\sum_{t=1}^{N_{\text{test}}}\sum_{d=1}^{D}\big(\hat z_{t,d}-z^{\text{true}}_{t,d}\big)^2,\qquad
		\mathrm{MAE}=\frac{1}{N_{\text{test}}D}\sum_{t=1}^{N_{\text{test}}}\sum_{d=1}^{D}\big|\hat z_{t,d}-z^{\text{true}}_{t,d}\big|.
		\label{eq:mse_mae}
	\end{equation}
	\item ALL of the true values under the marginal posteriors:
	\begin{equation}
		\mathrm{ALL}=\frac{1}{N_{\text{test}}D}\sum_{t=1}^{N_{\text{test}}}\sum_{d=1}^{D}\log q_{t,d}\big(z^{\text{true}}_{t,d}\big).
		\label{eq:all}
	\end{equation}
	\item Coverage: fraction of the true values that lie inside the central credible interval of level $p$, i.e., between the $(1-p)/2$ and $(1+p)/2$ quantiles of the marginal posterior:
	\begin{equation}
		C(p)=\frac{1}{N_{\text{test}}D}\sum_{t=1}^{N_{\text{test}}}\sum_{d=1}^{D}\mathbb{I}\Big[\frac{1-p}{2}\le F_{t,d}\big(z^{\text{true}}_{t,d}\big)\le\frac{1+p}{2}\Big],
		\label{eq:coverage}
	\end{equation}
	where $\mathbb{I}[\cdot]$ equals 1 if the condition holds, and 0 otherwise. The 95\% coverage is $C(0.95)$. If the posterior is calibrated, $C(p)\approx p$; a lower value (undercoverage) means that the true values fall outside the intervals more often than their level indicates.
	\item MACE over $n_l=19$ levels $p_l=0.05,0.10,\dots,0.95$:
	\begin{equation}
		\mathrm{MACE}=\frac{1}{n_l}\sum_{l=1}^{n_l}\big|C(p_l)-p_l\big|.
		\label{eq:mace}
	\end{equation}
\end{itemize}

\section{Sensitivity to \texorpdfstring{$\gamma$}{gamma}}
\label{app:gamma}
\begin{figure}[htbp]
	\centering
	\includegraphics[width=\textwidth]{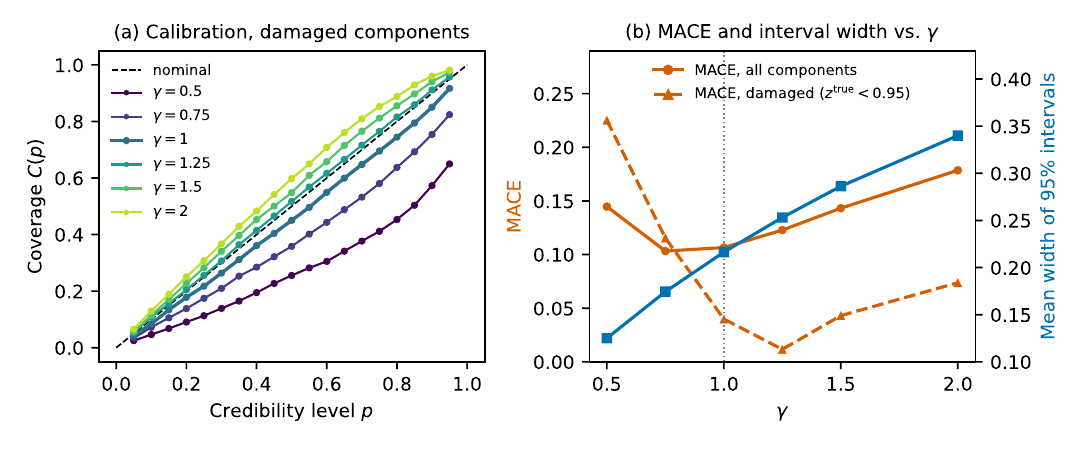}
	\caption{Sensitivity to $\gamma$ on the validation dataset: (a) coverage $C(p)$ of the damaged components; (b) MACE of all and of the damaged components, and mean width of the 95\% credible intervals.}
	\label{fig:gamma}
\end{figure}

For $\gamma\neq1$, the minimizer of Eq.~\eqref{eq:loss} is the variational approximation of the tempered posterior $p_\gamma(\mathbf{z}|\mathbf{m})\propto p(\mathbf{m}|\mathbf{z})^{1/\gamma^2}\,p(\mathbf{z})$.
A value $\gamma<1$ narrows the posterior (overconfident intervals), and $\gamma>1$ widens it (conservative intervals).
To document this effect, we retrained the model with $\gamma\in\{0.5,0.75,1,1.25,1.5,2\}$, with all the other settings unchanged.
Table~\ref{tab:gamma} reports the metrics of Appendix~\ref{sec:metrics} on the validation dataset, for all the components and for the damaged ones ($z^{\text{true}}<0.95$).
Fig.~\ref{fig:gamma} shows the calibration of the damaged components and the MACE.

\begin{table}[htbp]
	\centering\small
	\caption{Sensitivity to $\gamma$ on the validation dataset (2,400 observations). $W_{95}$: mean width of the 95\% credible intervals.}
	\label{tab:gamma}
	\footnotesize
	\setlength{\tabcolsep}{4pt}
	\begin{tabular}{ccccccccc}
		\toprule
		& \multicolumn{6}{c}{All components} & \multicolumn{2}{c}{Damaged components} \\
		\cmidrule(lr){2-7}\cmidrule(lr){8-9}
		$\gamma$ & MSE & MAE & ALL & $C(0.95)$ & MACE & $W_{95}$ & $C(0.95)$ & MACE \\
		\midrule
		0.5 & 0.00493 & 0.0479 & 1.222 & 0.766 & 0.145 & 0.125 & 0.650 & 0.225 \\
		0.75 & 0.00511 & 0.0530 & 1.689 & 0.859 & 0.103 & 0.174 & 0.824 & 0.115 \\
		\textbf{1} & \textbf{0.00584} & \textbf{0.0607} & \textbf{1.715} & \textbf{0.895} & \textbf{0.107} & \textbf{0.216} & \textbf{0.917} & \textbf{0.040} \\
		1.25 & 0.00696 & 0.0695 & 1.658 & 0.910 & 0.123 & 0.253 & 0.957 & 0.012 \\
		1.5 & 0.00839 & 0.0787 & 1.580 & 0.913 & 0.143 & 0.286 & 0.974 & 0.043 \\
		2 & 0.01167 & 0.0963 & 1.430 & 0.909 & 0.179 & 0.340 & 0.981 & 0.074 \\
		\bottomrule
	\end{tabular}
\end{table}

The results show the expected effect of $\gamma$:
\begin{itemize}
	\item The width of the credible intervals grows with $\gamma$, from 0.125 for $\gamma=0.5$ to 0.340 for $\gamma=2$.
	\item For $\gamma<1$, the posterior is overconfident: the 95\% coverage of the damaged components drops to 0.824 ($\gamma=0.75$) and 0.650 ($\gamma=0.5$). The point estimate improves (MAE of 0.048 for $\gamma=0.5$, against 0.061 for $\gamma=1$), mainly for the healthy components, whose narrower posteriors stay closer to the upper bound $z=1$.
	\item For $\gamma>1$, the intervals of the damaged components are conservative (coverage of 0.957--0.981), and the point estimate deteriorates (MAE up to 0.096).
\end{itemize}
With $\gamma=1$, the model yields the highest ALL, which rewards posteriors that are both accurate and well calibrated, and the posteriors of the damaged components are close to calibrated ($C(0.95)=0.917$, MACE of 0.040).
The slight undercoverage is typical of variational inference, which tends to underestimate the spread of the posterior.
For the healthy components, the 95\% coverage remains about 0.89 for every $\gamma\geq1$: about 11\% of their true values lie above the upper limit of the central 95\% interval, very close to the bound $z=1$.
This is an effect of using central intervals near the bound of the support, not of $\gamma$.
Therefore, the sensitivity analysis supports the value $\gamma=1$, fixed a priori, which provides the best balance between accuracy and calibration.

\section{Effect of the number of identified modes}
\label{sec:modes}
The case study already represents a situation of limited information: the parameterization used for training contains 10 elements (50 in the test dataset), and the seven stiffness factors are identified from five sensors and the first five modes. To analyze a situation with even less information, we have trained the model with only the first three modes, with the same sensors, damage scenarios, noise levels, likelihood calibration and training settings. Table~\ref{tab:modes} compares both setups on the test dataset.

\begin{table}[htbp]
	\centering
	\caption{Performance metrics on the test dataset with five and three identified modes. $W_{95}$: mean width of the 95\% credible intervals.}
	\label{tab:modes}
	\begin{tabular}{lcccccc}
		\toprule
		Identified modes & MSE & MAE & ALL & $C(0.95)$ & MACE & $W_{95}$ \\
		\midrule
		Five (case study) & 0.00570 & 0.0601 & 1.723 & 0.900 & 0.101 & 0.216 \\
		Three & 0.00818 & 0.0736 & 1.612 & 0.906 & 0.121 & 0.260 \\
		\bottomrule
	\end{tabular}
\end{table}

\begin{figure*}[htbp]
	\centering
	\includegraphics[width=\linewidth]{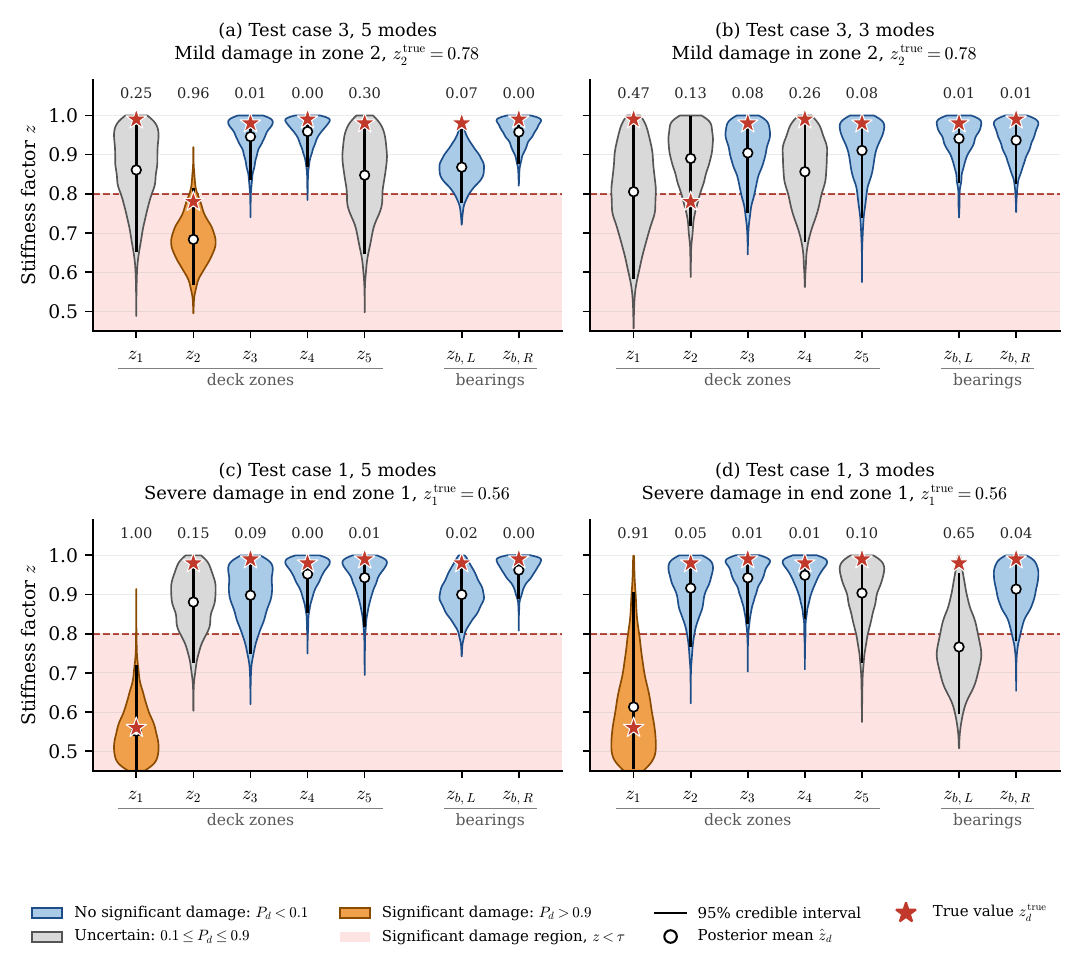}
	\caption{Posteriors of the stiffness factors (violin plots) of two scenarios of the test dataset identified with five modes (a, c) and three modes (b, d): (a, b) mild damage in zone 2; (c, d) severe damage in end zone 1. Colors and numbers as in Fig.~\ref{fig:cases_well}.}
	\label{fig:modes}
\end{figure*}

\begin{figure*}[htbp]
	\centering
	\includegraphics[width=\linewidth]{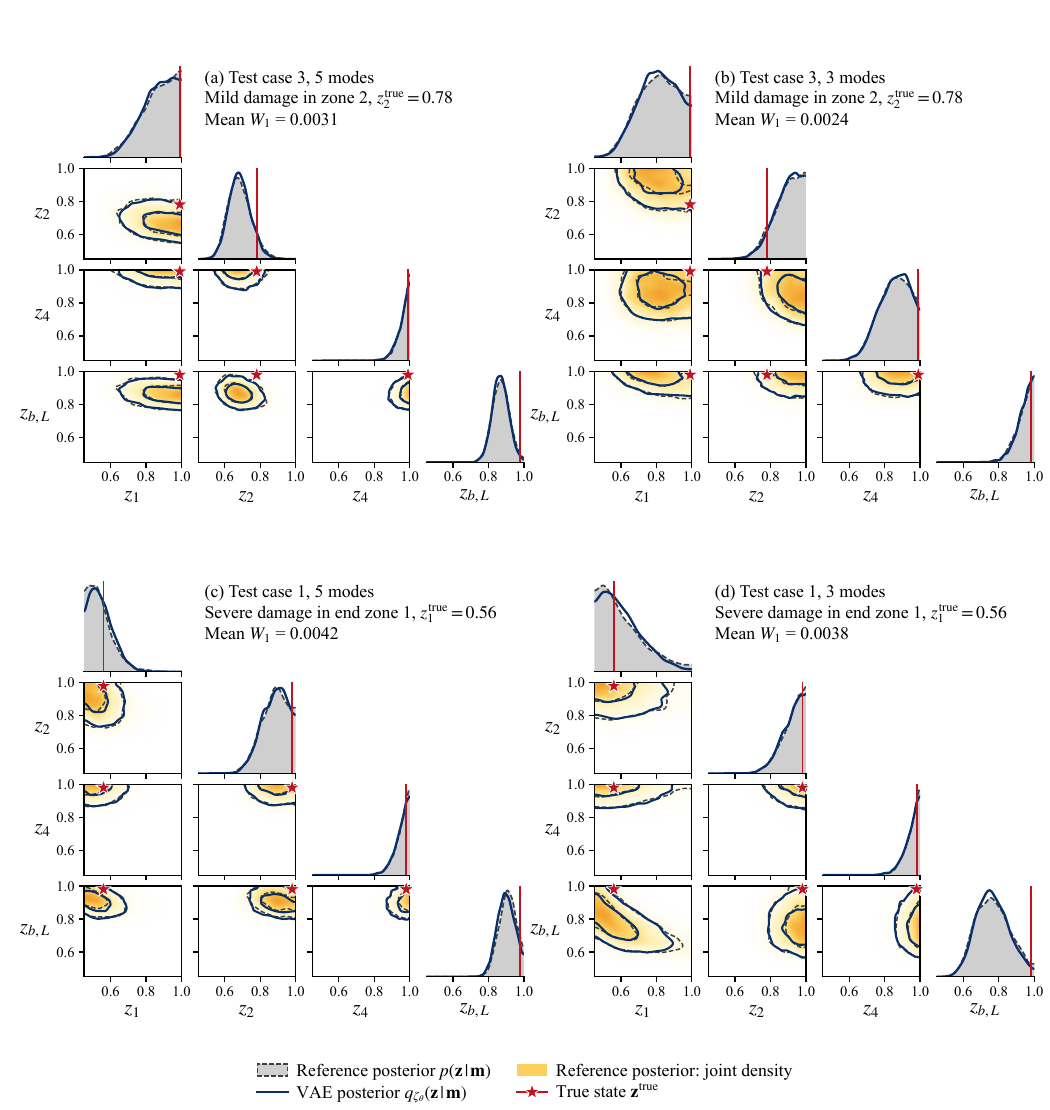}
	\caption{Corner plots of the VAE posterior and the exact (reference) posterior of the two scenarios of Fig.~\ref{fig:modes}, identified with five modes (a, c) and three modes (b, d), in the stiffness factors where the damage is located or spread: (a, b) mild damage in zone 2; (c, d) severe damage in end zone 1.}
	\label{fig:modes_corner}
\end{figure*}

With three modes, the measured data are less sensitive to the stiffness of each component, so the point estimates are less accurate (MAE of 0.074 against 0.060). However, the posteriors become wider (20\% larger mean width of the 95\% credible intervals), and the 95\% coverage remains the same (0.906 against 0.900): the VAE handles the loss of information by increasing the uncertainty of the estimates, instead of providing wrongly confident ones. Accordingly, the share of uncertain verdicts (Section~\ref{sec:detection}) increases, e.g., from 25\% to 36\% for the healthy components and from 23\% to 44\% for the losses of stiffness larger than 30\%, whereas the wrong confident verdicts remain rare (0.3\% of false positives and 1.6\% of false negatives). Fig.~\ref{fig:modes} shows two scenarios of the test dataset with both setups. For a mild damage in zone 2 ($z_2^{\text{true}}=0.78$), zone 2 is classified as significant damage with five modes ($P_d=0.96$), whereas with three modes the probability of damage spreads over zones 1, 2 and 4, which are all uncertain. For a severe damage in the end zone 1 ($z_1^{\text{true}}=0.56$), the damage is classified as significant with both setups, but with three modes the adjacent left bearing becomes uncertain ($P_d=0.65$), since a loss of stiffness of an end zone and a softer bearing have similar effects on the first three modes. Fig.~\ref{fig:modes_corner} compares the VAE posteriors of both scenarios with the exact posteriors computed with the same measurements. With both setups, the VAE closely matches the exact posterior (mean Wasserstein-1 distance between the marginals below 0.005). With three modes, the exact posterior itself becomes wider: in the mild damage scenario, it spreads the loss of stiffness over zones 1, 2 and 4, and in the severe damage scenario, it shows a strong negative correlation between $z_1$ and $z_{b,L}$, since the first three modes cannot separate a loss of stiffness of the end zone from a softer bearing. Hence, the wider posteriors are caused by the lower information content of the measured data, not by the approximation of the VAE. The same holds for the 11 benchmark scenarios of the test dataset (mean Wasserstein-1 distance of 0.002--0.007): in the transition to a severely ill-posed problem, the VAE follows the exact posterior.

\section*{Data Availability Statement}
Some or all data, models, or code generated or used during the study are available in a repository online: the code and the synthetic datasets can be found at \url{https://github.com/Mathmode/RCopulaVAE_eigenproblem_beam}.

\section*{Acknowledgements}
This work has received funding from the following research projects/grants/institutions:
The Basque Government, through the postdoctoral grant $\text{POS}\_2025\_1\_0098$; the BERC 2022-2025 program; the ELKARTEK program with projects RUL-ET (KK-2024/00086), SEGURH2 (KK-2024/00068); and the IKUR-HPC$\&$AI program (HPCAI7.OceaNNic).
The European Union's Horizon Europe research and innovation programme under the Marie Sk\l{}odowska-Curie Action MSCA-and-citizens 101162248 -ORE4Citizens, the MSCA-DN-101119556 (IN-DEEP), and FUTURAL (101083958) project (which supports research on the assessment of beam-type bridges in rural areas using artificial intelligence), and SAFEBIDE (ZL-2024/00272).
The project PID2023-146678OB-I00 funded by MICIU/AEI/10.13039/501100011033 and by FEDER, EU.
The BCAM Severo Ochoa accreditation of excellence CEX2021-001142-S funded by MICIU/AEI/10.13039/501100011033.
The consolidated Research Group MATHMODE (IT1866-26) of the UPV/EHU, given by the Department of Education of the Basque Government.
BCAM-IKUR-UPV/EHU, funded by the Basque Government IKUR Strategy and by the European Union NextGenerationEU/PRTR.

\section*{Declaration of generative AI use}
During the preparation of this work, the authors used Gemini Pro to improve code quality and efficiency and enhance the writing process. After using this tool, the authors carefully reviewed and edited the content as needed and take full responsibility for the content of the published article.

\end{document}